\documentclass[10pt,twocolumn,letterpaper]{article}

\usepackage{cvpr}              

\usepackage{graphicx}
\usepackage{amsmath}
\usepackage{amssymb}
\usepackage{booktabs}

\usepackage{multirow}
\usepackage{tabularx}

\usepackage[accsupp]{axessibility}  

\usepackage[pagebackref,breaklinks,colorlinks]{hyperref}

\usepackage[capitalize]{cleveref}
\crefname{section}{Sec.}{Secs.}
\Crefname{section}{Section}{Sections}
\Crefname{table}{Table}{Tables}
\crefname{table}{Tab.}{Tabs.}

\def\cvprPaperID{4166} 
\def\confName{CVPR}
\def\confYear{2023}

\begin{document}

\title{DA-DETR: Domain Adaptive Detection Transformer with Information Fusion}

\author{Jingyi Zhang\textsuperscript{\rm 1,}\thanks{Equal contribution, \{jingyi.zhang, jiaxing.huang\}@ntu.edu.sg.} \  Jiaxing Huang\textsuperscript{\rm 1,}$^*$ \  Zhipeng Luo\textsuperscript{\rm 1,2} \  Gongjie Zhang\textsuperscript{\rm 1,3} \  Xiaoqin Zhang\textsuperscript{\rm 4} \  Shijian Lu\textsuperscript{\rm 1,}\thanks{Corresponding author, shijian.lu@ntu.edu.sg.} 
\\
$^1$ S-lab, Nanyang Technological University \ \  $^2$ SenseTime Research \\
 $^3$ Black Sesame Technologies \ \ $^4$ Wenzhou University
\\
}
\maketitle

\begin{abstract}
The recent detection transformer (DETR) simplifies the object detection pipeline by removing hand-crafted designs and hyperparameters as employed in conventional two-stage object detectors. However, how to leverage the simple yet effective DETR architecture in domain adaptive object detection is largely neglected. Inspired by the unique DETR attention mechanisms, we design DA-DETR, a domain adaptive object detection transformer that introduces information fusion for effective transfer from a labeled source domain to an unlabeled target domain. DA-DETR introduces a novel CNN-Transformer Blender (CTBlender) that fuses the CNN features and Transformer features ingeniously for effective feature alignment and knowledge transfer across domains.
Specifically, CTBlender employs the Transformer features to modulate the CNN features across multiple scales where the high-level semantic information and the low-level spatial information are fused for accurate object identification and localization. Extensive experiments show that DA-DETR achieves superior detection performance consistently across multiple widely adopted domain adaptation benchmarks.

\end{abstract}

\section{Introduction}
\label{sec:intro}

Object detection aims to predict a bounding box and a class label for interested objects in images and it has been a longstanding challenge in the computer vision research. Most existing work adopts a two-stage detection pipeline that involves heuristic anchor designs, complicated post-processing such as non-maximum suppression (NMS), etc. The recent detection transformer (DETR)~\cite{carion2020detr} has attracted increasing attention which greatly simplifies the two-stage detection pipeline by removing hand-crafted anchors~\cite{girshick2014rcnn,girshick2015fastrcnn, ren2015fasterrcnn} and NMS~\cite{girshick2014rcnn, girshick2015fastrcnn, ren2015fasterrcnn}. Despite its great detection performance under a fully supervised setup, how to leverage the simple yet effective DETR architecture in domain adaptive object detection is largely neglected.

\begin{figure}[t]
\centering
\includegraphics[width=1.0\linewidth]{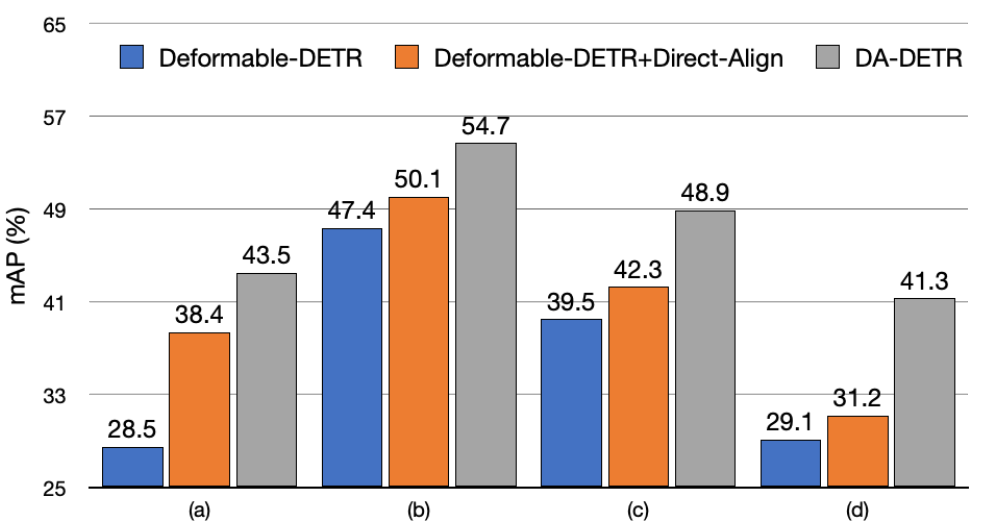}
\caption{
The vanilla \textit{Deformable-DETR}~\cite{zhu2020deformable} trained with labeled source data cannot handle target data well due to cross-domain shift. The introduction of adversarial feature alignment in \textit{Deformable-DETR + Direct-align}~\cite{ganin2015grl} improves the detection clearly. The proposed \textit{DA-DETR} fuses CNN features and transformer features ingeniously which achieves superior unsupervised domain adaptation consistently across four widely adopted benchmarks including Cityscapes $\rightarrow$ Foggy cityscapes in (a), SIM 10k $\rightarrow$ Cityscapes in (b), KITTI $\rightarrow$ Cityscapes in (c) and PASCAL VOC $\rightarrow$ Clipart1k in (d).
}
\label{fig:intro}
\end{figure}

Different from the conventional CNN-based detection architectures such as Faster RCNN~\cite{ren2015fasterrcnn}, DETR has a CNN backbone followed by a transformer head consisting of an encoder-decoder structure. The CNN backbone and the transformer head learn different types of features~\cite{d2021convit,yuan2021tokens,raghu2021vision} - the former largely captures low-level localization features ($e.g.$, edges and lines around object boundaries) while the latter largely captures global inter-pixel relationship and high-level semantic features. At the other end, many prior studies show that fusing different types of features often is often helpful in various visual recognition tasks~\cite{dai2021attentional,cheng2017locality}. Hence, it is very meaningful to investigate how to fuse the two types of DETR features to address the domain adaptive object detection challenge effectively.

We design DA-DETR, a simple yet effective Domain Adaptive DETR that introduces information fusion into the DETR architecture for effective domain adaptive object detection. The core design is a CNN-Transformer Blender (CTBlender) that employs the high-level semantic features in the Transformer head to conditionally modulate the low-level localization features in the CNN backbone. CTBlender consists of two sequential fusion components, including split-merge fusion (SMF) that fuses CNN and Transformer features within an image and scale aggregation fusion (SAF) that fuses the SMF features across multiple feature scales. Different from the existing weight-and-sum fusion~\cite{dai2021attentional,cheng2017locality}, 
SMF first splits CNN features into multiple groups with different semantic information as captured by the Transformer head and then merges them with channel shuffling for effective information communication among different groups.
The SMF features of each scale are then aggregated by SAF for fusing both semantic and localization information across multiple feature scales. Hence, CTBlender captures both semantic and localization features ingeniously which enables comprehensive and effective inter-domain feature alignment with a single discriminator.

The main contributions of this work can be summarized in three aspects. 
\textit{First}, we propose DA-DETR, a simple yet effective domain adaptive detection transformer that introduces information fusion for effective domain adaptive object detection.
To the best of our knowledge, this is the first work that explores information fusion for domain adaptive object detection.
\textit{Second}, we design a CNN-Transformer Blender that fuses the CNN features and Transformer features ingeniously for effective feature alignment and knowledge transfer across domains.
\textit{Third}, extensive experiments show that DA-DETR achieves superior object detection over multiple widely studied domain adaptation benchmarks as compared with the state-of-the-art as shown in Fig.~\ref{fig:intro}.

\section{Related Work}
\label{related_work}

\noindent \textbf{Transformers}~\cite{vaswani2017transformer} have achieved great success in various neural language processing (NLP) tasks~\cite{devlin2018bert,radford2018improving,radford2019language,brown2020language,liu2019roberta} due to their computational efficiency and scalability. Inspired by the success of transformers in NLP, several studies~\cite{dosovitskiy2020vit,zheng2021setr,xie2021segformer,cheng2021maskformer,carion2020detr,zhu2020deformable,dai2021up,zhao2021point,dong2021solq,zhang2022towards,zhang2022meta} attempt to adapt transformers to computer vision tasks. For example, ViT~\cite{dosovitskiy2020vit} adopts transformer for image classification, which splits each image into patches and treats them as input tokens to transformers. SETR~\cite{zheng2021setr} extends ViT to semantic segmentation by introducing multiple decoders designed for pixel-wise segmentation. 
For object detection, different from CNN-based detector, DETR~\cite{carion2020detr} treats detection as a set prediction task~\cite{rezatofighi2017deepsetnet}, which eliminates the dependence on various heuristic and hand-crafted designs such as anchor generation, ROI pooling and non-maximum suppression.
Existing vision transformers achieve very promising performance in supervised learning. However, how to adapt and generalize them to unsupervised domain adaptation tasks has been largely neglected. In this work, we investigate domain adaptive detection transformers in an unsupervised manner.

\noindent \textbf{Unsupervised Domain Adaptation (UDA)} 
has been studied extensively in recent years, largely for alleviating data annotation constraint in deep network training in various visual recognition tasks ~\cite{ganin2015grl,chen2018wild,zou2018self_seg,vu2019advent,saito2018maximum,pinheiro2018unsupervised,zou2019confidence,yang2020fda,huang2021model,huang2020contextual,zhang2021proda,huang2021cross,guan2021scale,xing2022domain,huang2022category,zhang2022spectral,luo2022domain,luo2021unsupervised}.
For object detection, the target of UDA is to mitigate the domain gap between a source domain and a target domain, so that the source data can be employed to train better detectors for target data. Most existing domain adaptive detectors \cite{chen2018wild,saito2019strong,2020coarse2fine,li2020SAP,shen2019gradientdetach,xu2020category,zhu2019selective,chen2020harmonizing,vs2021mega,zhang2021rpn} adopt CNN-based detector ($i.e.$, Faster R-CNN) and achieve UDA via adversarial learning~\cite{chen2018wild,saito2019strong,2020coarse2fine,li2020SAP,shen2019gradientdetach,xu2020category,he2019MAF}, image translation~\cite{arruda2019daynight,kim2019diversify,inoue2018weakly,shan2019pixelgan,yu2019self-training,arruda2019daynight,huang2021fsdr,lin2019on-road} and self-training~\cite{yu2019self-training,huang2021model}. 
However, little research~\cite{wang2021exploring,yu2022mttrans} is conducted on how to adopt DETR in domain adaptive detection tasks, $e.g.$, SFA~\cite{wang2021exploring} tackles the domain adaptive object detection via query-based feature alignment and token-wise feature alignment.
Differently, we introduce the information fusion idea into the DETR architecture for effective domain adaptive object detection.

\noindent \textbf{Feature Fusion} 
is often helpful in various visual recognition tasks. For example, ResNet~\cite{he2016resnet} fuses features of different network layers with skip connections which achieves obvious performance gains along with increased network depth. Feature Pyramid Network (FPN)~\cite{lin2017feature} fuses features of different scales to build high-level semantic feature maps, and it usually improves the object detection with clear margins in various detection tasks. 
Some work~\cite{dai2021attentional,cheng2017locality} instead achieves feature fusion with channel-wise attention~\cite{hu2018squeezenet,qilong2020channelatt} and spatial-wise attention~\cite{wang2018sptailatt,woo2018cbam,cao2019gcnet,li2019sge}.
For example, AFF~\cite{dai2021attentional} presents a multi-scale channel attention module for better fusing features from different layers and scales that capture different types of semantics. LS-DeconvNet~\cite{cheng2017locality} introduces a gated fusion layer to fuse RGB and depth features effectively.
We explore information fusion for domain adaptive detection transformer, and design a CNN-Transformer Blender for fusing CNN features and Transformer features for effective domain adaptive object detection. Different from~\cite{dai2021attentional, cheng2017locality} that adopts a weight-and-sum strategy, our CNN-Transformer Blender employs the Transformer features to modulate the CNN features for more effective adversarial feature alignment.

\begin{figure*}[!ht]
\centering
\includegraphics[width=0.9\linewidth]{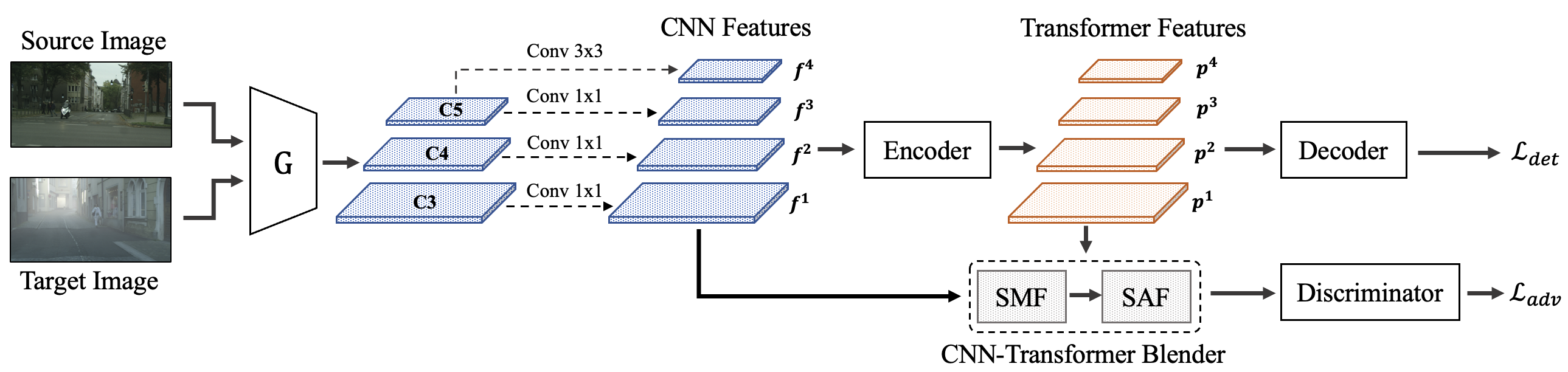} 
\caption{
{Overview of the proposed DA-DETR: 
the proposed DA-DETR consists of a base detector (including a backbone $G$ and a transformer encoder-decoder), a discriminator and a CNN-Transformer Blender (CTBlender). Given an input image from either source or target domain, the backbone $G$ first produces multi-scale CNN features $f^l$ ($ l=1,2,3,4$) and then feeds them to the transformer encoder to obtain Transformer features $p^l$ ($ l=1,2,3,4$). For \textit{supervised learning}, the Transformer features generated by the source images are further fed to the decoder to compute supervised detection loss $\mathcal{L}_{det}$ with the corresponding ground truth.
For \textit{unsupervised learning}, CTBlender takes $f^l$ and $p^l$ as inputs for feature fusion. Finally, the output of CTBlender is fed to the discriminator for computing an adversarial loss $\mathcal{L}_{adv}$ which drives adversarial alignment of source and target features.}
}
\label{fig:stru}
\end{figure*}

\section{Preliminaries of Detection Transformer}
\label{preliminaries}

DETR~\cite{carion2020detr} consists of a CNN backbone~\cite{he2016resnet} to extract features, an encoder-decoder transformer and a simple feed forward network (FFN) to make final detection prediction.
Given an image $x$, the CNN backbone $G$ first generates feature $f$ and then reshapes $f$ to a vector. 
The encoder-decoder in DETR follows the standard architecture of the transformer~\cite{vaswani2017transformer}, which consists of multiple multi-head self-attention modules that are defined by:
\begin{equation}
\label{multi-h}
\begin{split}
MSA(z_q, f) = \sum_{h=1}^H {P}_H [\sum_k SA_{hqk} \cdot {{P}_H}^\prime f_k],
\end{split}
\end{equation}
where $MSA(\cdot)$ consists of $H$ single attention heads, $z_q$ and $f_k$ denotes representation features of query element and key element, ${P}_H \in \mathbb{R}^{d \times d_h}$ and ${{P}_H}^\prime \in \mathbb{R}^{d \times d_h}$ are learnable projection weights ($d_h = d / H$, where $d$ is the dimension of $f$). 
Each self-attention weight $SA_{hqk}$ is a type of scaled dot-product attention, which maps a query and a set of key-value pairs into an output:
\begin{equation}
\label{single_att_1}
\begin{split}
SA_{hqk} \propto \exp{(\frac{z_q^T U_m^T V_m f_c}{\sqrt{d_h}})},
\end{split}
\end{equation}
where $U_m, V_m \in \mathbb{R}^{d_h \times d}$ are also learnable weights.

We adopt Deformable-DETR~\cite{zhu2020deformable} as the base detector. Different from the conventional DETR~\cite{carion2020detr}, Deformable-DETR replaces the normal attention with the deformable attention which improves the convergence speed greatly:
\begin{equation}
\label{deform-att}
\begin{split}
Defor&mableMSA(z_q, p_q, f) \\ &=  \sum_{h=1}^H {P}_H [\sum_k SA_{hqk} \cdot {{P}_H}^\prime f (p_q + \delta p_{hqk})],
\end{split}
\end{equation}
where $\delta p_{hqk}$ and $SA_{hqk}$ denote the sampling offset and attention weight of the k-th sampling point in the m-th attention head, respectively.
Such sampling design significantly mitigates the slow convergence and high complexity issues of DETR~\cite{carion2020detr}. 
In addition, Deformable-DETR is extended to aggregating multi-scale features as shown in Fig.~\ref{fig:stru}.
The multi-scale feature maps $f^l$ ($l=1,2,3,4$) are extracted from the output of Block C3-C5 in the ResNet backbone~\cite{he2016resnet}.
More specifically, $f^1$, $f^2$ and $f^3$ are extracted from the output feature maps of Block C3-C4 via a 1×1 convolution. The lowest resolution feature map, $i.e.$, $f^4$, is extracted by 3×3 stride 2 convolution on the output feature maps of Block C5. 
Such multi-scale design enables attention to capture relationships among different-scale features effectively.

\begin{figure*}[!ht]
\centering
\includegraphics[width=.88\linewidth]{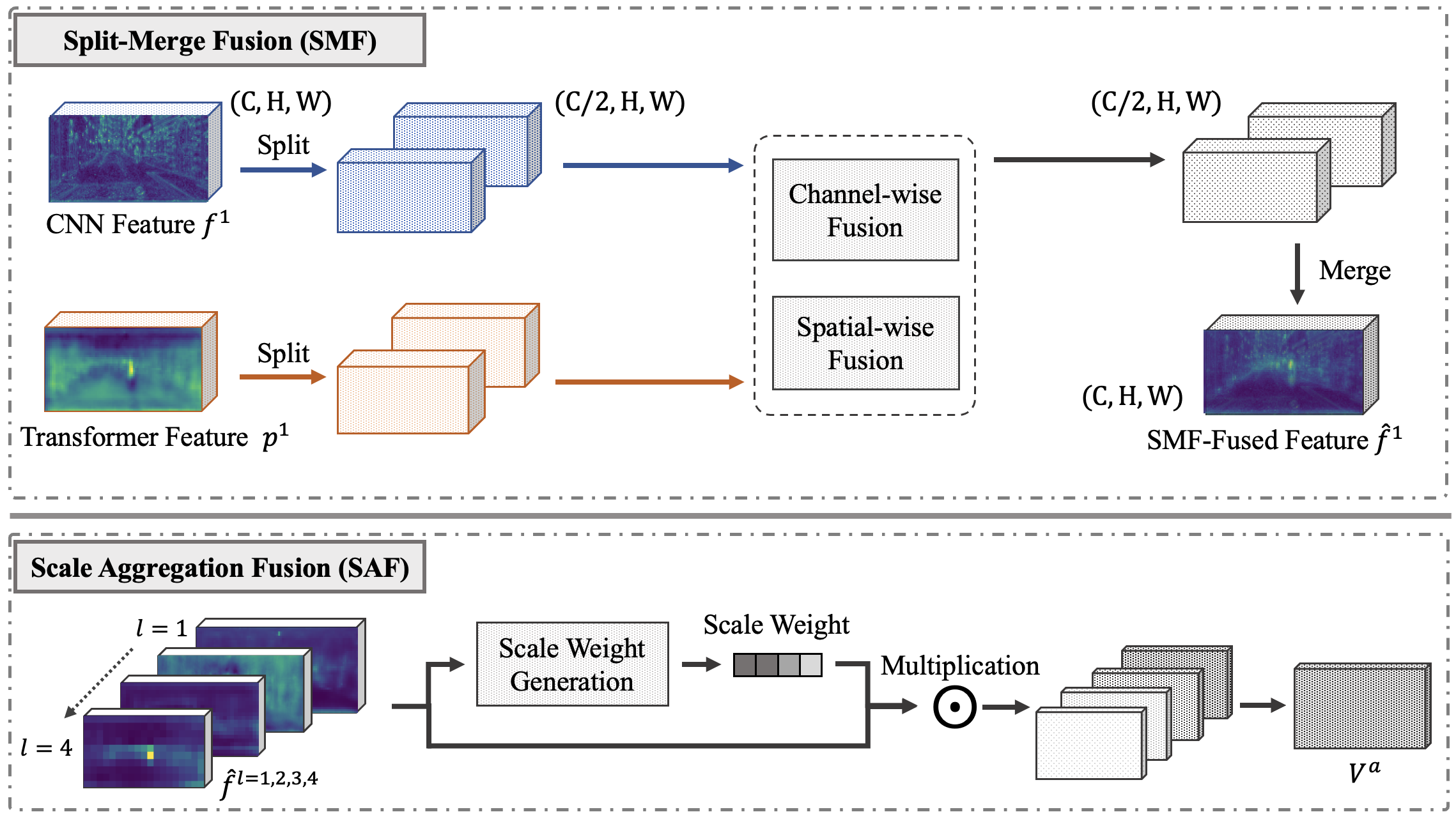}
\caption{
{Overview of the proposed CNN-Transformer Blender (CTBlender).
CTBlender consists of split-merge fusion (SMF) and scale aggregation fusion (SAF) as illustrated. In SMF, Transformer features of all four scales are adopted to modulate the CNN features. Take the first level $f^{1}$ and $p^{1}$ as an example. 
The Transformer feature $p^{1}$ and the CNN feature $f^{1}$ are divided into $K$ groups ($e.g.$, $K=2$), and further fed to SMF to perform spatial-wise fusion and channel-wise fusion, respectively.
The fused group features are then merged to generate the final fused features ${\hat{f}}^{1}$ for the first scale. In SAF, $\hat{f}^l$ (${l=1,2,3,4}$) are aggregated with different scale-wise weights to generate the final feature $V^a$.}
}
\label{fig:ham}
\end{figure*}

\section{Method}
\label{method}

\subsection{Task Definition}
\label{3_1}

The work focuses on the problem of unsupervised domain adaptation (UDA) in object detection. It involves a source domain $\mathcal{D}_s$ and a target domain $\mathcal{D}_t$, where $\mathcal{D}_s=\left\{\left(x_{s}^{{i}}, y_{s}^{{i}}\right)\right\}_{i=1}^{N_{{s}}}$ is fully labeled, and $y_{s}^{{i}}$ represents the labels of the sample image $x_{s}^{{i}}$. The goal is to train a detection transformer that well performs on unlabeled target-domain data $x_{t}^i$. The baseline model is trained with the labeled source data ($i.e.$, $\mathcal{D}_s$) only:

\begin{equation}
\label{eq1}
\mathcal{L}_{det}=l(T(G(x_s)), y_s),
\end{equation}
where $G$ denotes backbone, $T$ denotes transformer encoder-decoder and $l\left(\cdot \right)$ denotes the supervised detection loss that consists of a matching cost and a Hungarian loss~\cite{carion2020detr,zhu2020deformable} for object category and object box predictions.

\subsection{Framework Overview}
\label{3_2}

As shown in Fig.~\ref{fig:stru}, the proposed DA-DETR consists of a base detector (including a backbone $G$ and a transformer encoder-decoder $T$), a discriminator $C_{d}$ and a CNN-Transformer Blender (CTBlender). We adopt the deformable-DETR~\cite{zhu2020deformable} as the base detector, where $G$ extracts features from the input images and $T$ predicts a set of bounding boxes and pre-defined semantic categories according to the extracted features. CTBlender consists of two sub-modules including a \textit{split-merge fusion (SMF)} and a \textit{scale aggregation fusion (SAF)} as in Fig.~\ref{fig:ham}. Taking the CNN features from $G$ and the Transformer features from the encoder $E$ as inputs, CTBlender fuses the positional and semantic information in Transformer features and the localization information in CNN features for comprehensive and effective feature alignment across domains.

Given an input image from either source or target domain, the backbone $G$ will first produce multi-scale features $f^l$ ($ l=1,2,3,4$) and then feeds them to the transformer encoder to obtain Transformer features $p^l$ ($ l=1,2,3,4$). For \textit{supervised learning}, the Transformer features generated by source images $x_s\in\mathcal{D}_s$ are further fed to decoder to predict a set of bounding boxes and pre-defined semantic categories, which will be used to calculate a detection loss $\mathcal{L}_{det}$ under the supervision of the corresponding ground-truth label $y_s \in \mathcal{D}_s$.
For \textit{unsupervised learning}, CTBlender takes $f^l$ ($ l=1,2,3,4$) and $p^l$ ($ l=1,2,3,4$) generated by both source and target images ($i.e.$, $x_s\in\mathcal{D}_s$ and $x_t\in\mathcal{D}_t$) as inputs. 
Finally, the output of CTBlender is fed to the discriminator $C_d$ to compute an adversarial loss $\mathcal{L}_{adv}$ for inter-domain feature alignment.
The overall network is optimized by the adversarial loss $\mathcal{L}_{adv}$ and the detection loss $\mathcal{L}_{det}$.

\subsection{{CNN-Transformer Blender}}
\label{ham}

One key component in DA-DETR is a CNN-Transformer Blender (CTBlender) that fuses different features for effective domain alignment.
CTBlender takes multi-scale CNN features  and the corresponding multi-scale Transformer features as the input, where the semantic and positional information in the Transformer features are fused with the localization information in the CNN features via split-merge fusion (SMF). The SMF-fused features are then aggregated across multiple scales via scale aggregation fusion (SAF).

\noindent\textbf{Split-Merge Fusion.}
In SMF, the rich semantic and positional information in the the multi-scale Transformer features $p= \left\{ p^l\right\}_{l=1}^{L}$ are exploited to fuse with multi-scale CNN features $f= \left\{ f^{l}\right\}_{l=1}^{L}$ for adversarial feature alignment across domains.
As SMF operations at each feature scale are the same, we take the first scale $l=1$ as an example to illustrate how we perform split-merge fusion.

Inspired by the split-fuse-merge in~\cite{zhang2020resnest}, SMF first splits CNN features into multiple groups and then fuses them with the Transformer features. After that, the fused features are merged with channel shuffling for effective information communication among different groups.
Given a Transformer feature $p^1 \in \mathbb{R}^{C \times H \times W}$ and a backbone CNN feature $f^1 \in \mathbb{R}^{C \times H \times W}$ ($C$, $H$, $W$ indicate the number of channel of feature map, and the height and the width of feature map, respectively), $p^1$ is first split into $K$ groups evenly along channels, $i.e.$, $\left\{p^1_k\right\}^{K}_{k=1} \in \mathbb{R}^{(C/K) \times H \times W}$, where each group captures different semantic information of the input image. 
The fusion in each group is then achieved via spatial-wise fusion and channel-wise fusion, respectively.

For the spatial-wise fusion, the split Transformer features are firstly fed into a normalization layer and then re-weighted by a learnable weight map and a learnable bias map:

\begin{equation}
\label{eq2}
\hat{p}_{ks}^{1}=f_s \left(w_{s} \cdot {GN}\left(p_{k}^1\right)+b_{s}\right),
\end{equation}
where $f_s(\cdot)$ is an activation function that limits the input in the range of $[0, 1]$.

For the channel-wise fusion, the split Transformer feature is firstly compacted by the Global Average Pooling (GAP) and then re-weighted by a learnable weight vector and a learnable bias vector:

\begin{equation}
\label{eq4}
\hat{p}_{kc}^{1}=f_s\left(w_{c} \cdot {{GAP}(p_{k}^1)+b_{c}}\right),
\end{equation}
where $f_s(\cdot)$ is an activation function that limits the input to the range of $[0, 1]$. 

Similar to the operation for Transformer feature $p^1$, the CNN feature $f^1$ is also divided into $K$ groups along the channels, $i.e.$, $\left\{f^1_k\right\}^{K}_{k=1} \in \mathbb{R}^{(C/K) \times H \times W}$.

We further adopt shuffle operation to enable information communication across channels~\cite{ma2018shufflenet,zhang2018shufflenet}. Specifically, we first re-weight the split CNN feature by the corresponding re-weighted Transformer feature ($i.e.$, $\hat{p}_{ks}^{1}$ and $\hat{p}_{kc}^{1}$) to generate re-weighted split CNN feature $\hat{f_k^1}$.

Then we shuffle $\hat{f}_{k}^1$ along channels to enable information flow across channels for better feature fusion.
Lastly, we conduct the above operations $K$ times to generate $K$ shuffled features for each group, $i.e.$, $\left\{\hat{f}^1_k\right\}^{K}_{k=1} \in \mathbb{R}^{(C/K) \times H \times W}$. The shuffled features are concatenated to obtain the fused feature map $\hat{f}^1 \in \mathbb{R}^{C \times H \times W}$:

\begin{equation}
\label{eq6}
{\hat{f}}^1= f_c \left({\hat{f}_{1}^1},...,{\hat{f}_{k}^1},...,{\hat{f}_{K}^1}\right),
\end{equation}
where similar operations are conducted to get the results of all levels ${\hat{f}}= \left\{ {\hat{f}}^{l}\right\}^{L}_{l=1}$.

\noindent \textbf{Scale Aggregation Fusion.}
To explicitly perform feature fusion of different scales, we design a scale aggregation fusion (SAF) to aggregate features ${\hat{f}}$ with different scale weights as illustrated in the bottom part of Fig.~\ref{fig:ham}.

Specifically, we compact each scale of feature ${\hat{f}} = \left\{ {\hat{f}}^{l}\right\}^{L}_{l=1}$ into a channel-wise vector ${u}= \left\{ {u}^{l}\right\}^{L}_{l=1} \in \mathbb{R}^{C \times 1 \times 1}$ via a Global Average Pooling (GAP) layer. The scale weights $\alpha_l$ are obtained from channel-wise vectors ${u}^l$. Firstly, the channel-wise vectors are merged together to obtain merged vector $u_{m}$ by an element-wise addition.

Then, a fully connected layer separates $u_{m}$ to $L$ scale-weight vectors $\alpha^l \in \mathbb{R}^{C \times 1 \times 1}$. Finally, $V^a$ is obtained by
\begin{equation}
\label{8}
V^a=\sum_{l=1}^{L} {\hat{f}^l} \cdot \alpha^l,
\end{equation}
where $V^a$ is a highly embedded feature that captures rich semantic information and localization information.

\begin{table}[t]
\centering
\footnotesize
\begin{tabular}{cccc|c}
\toprule
\multicolumn{1}{c|}{\multirow{2}{*}{Direct-align}} &\multicolumn{2}{c|}{+SMF} &\multicolumn{1}{c|}{\multirow{2}{*}{+SAF}} &\multicolumn{1}{c}{\multirow{2}{*}{mAP}}
\\
\cmidrule{2-3}
\multicolumn{1}{c|}{\multirow{1}{*}{}} &Shuffling & \multicolumn{1}{c|}{Splitting} &\multicolumn{1}{c|}{\multirow{1}{*}{}}&\multicolumn{1}{c}{\multirow{1}{*}{}}
\\
\midrule
& & & &28.5\\
\midrule
\multicolumn{1}{c}{\checkmark} & & & &38.4\\
\multicolumn{1}{c}{\checkmark} &\multicolumn{1}{c}{\checkmark} & & &41.4\\
\multicolumn{1}{c}{\checkmark} & & \multicolumn{1}{c}{\checkmark} & &41.8\\
\multicolumn{1}{c}{\checkmark} &\multicolumn{1}{c}{\checkmark} &\multicolumn{1}{c}{\checkmark} &  &42.3\\
\multicolumn{1}{c}{\checkmark} & & & \multicolumn{1}{c|}{\checkmark} &41.7\\
\multicolumn{1}{c}{\checkmark} & \multicolumn{1}{c}{\checkmark}& \multicolumn{1}{c}{\checkmark}& \multicolumn{1}{c|}{\checkmark}&\textbf{43.5}\\
\bottomrule
\end{tabular}
\caption{
Ablation study of DA-DETR over domain adaptation task Cityscapes $\rightarrow$ Foggy Cityscapes.
}
\label{ablation_1}
\end{table}

\subsection{Network Training}
\label{3_5}

The network is trained with two losses, $i.e.$, a supervised object detection loss $\mathcal{L}_{det}$ as defined in Eq.~\ref{eq1} and an adversarial alignment loss $\mathcal{L}_{adv}$ that is defined as follow:
\begin{equation}
\begin{array}{rl}
\label{9}
\mathcal{L}_{adv} = & \mathbb{E}_{(f,p) \in  \mathcal{D}_{s}} \log {C_d}\left(\mathcal{H}\left(f, p) \right) \right)\\
& + \mathbb{E}_{(f,p) \in \mathcal{D}_{t}} \log \left(1- {C_d}\left(\mathcal{H}\left(f, p \right) \right) \right),
\end{array}
\end{equation}
where $f = G\left(x\right)$ and $p = E\left(G \left(x \right)\right)$. $G$ denotes backbone; $E$ denotes transformer encoder; $\mathcal{H}$ denotes CNN-Transformer Blender (CTBlender) and $C_d$ denotes the discriminator. Both source images $x_s$ and target images $x_t$ are utilized to compute adversarial loss. 

In summary, the overall optimization objective of DA-DETR is formulated by
\begin{equation}
\label{10}
\max _{C_d} \min _{G, T, \mathcal{H}} \mathcal{L}_{\mathrm{det}}(G, T)-\lambda \mathcal{L}_{\mathrm{adv}}(\mathcal{H}, C_d),
\end{equation}
where $T$ denotes the transformer in DETR, $\lambda$ is the weight factor that balances the influences of $\mathcal{L}_{{det}}$ and $\mathcal{L}_{{adv}}$ in training. Note that we adopt a gradient reverse layer (GRL)~\cite{ganin2015grl} to enable the gradient of $\mathcal{L}_{{adv}}$ to be reversed before back-propagating to $\mathcal{H}$ from $C_d$.

\begin{table*}[!t]
\centering
\begin{footnotesize}
\begin{tabular}{c|c|cccccccc|c}
\toprule
\multicolumn{11}{c}{\textbf{Cityscapes $\rightarrow$ Foggy cityscapes}} \\
\midrule
Method & Backbone & person & rider & car & truck & bus & train & mcycle & bicycle & mAP \\
\midrule
Deformable-DETR~\cite{zhu2020deformable} & ResNet-50 & 37.7& 39.1& 44.2& 17.2& 26.8& 5.8 &21.6 &35.5 &28.5 \\
\midrule
DAF ~\cite{chen2018wild} & ResNet-50 & 48.2 & 48.8 & 61.5 & 22.6 & 43.1 & 20.2 & 30.3 & 42.1 & 39.6
\\
SWDA ~\cite{saito2019strong} & ResNet-50 & 49.0 & 49.0 & 61.4 & 23.9 & 43.1 & 22.9 & 31.0 & 45.2 & 40.7
\\
{SCL}~\cite{shen2019gradientdetach}& {ResNet-50} &{49.4}&{48.6}&{61.2}&{\textbf{27.2}}&{41.1}&{34.8}&{28.5}&{42.5}&{41.7}\\
{GPA}~\cite{xu2020graph}& {ResNet-50} & {49.5}&{46.7}&{58.6}&{26.4}&{42.2}&{32.3}&{29.1}&{41.8}&{40.8}\\ 
CRDA ~\cite{xu2020category} & ResNet-50 & 49.8 & 48.4 & 61.9 & 22.3 & 40.7 & 30.0 & 29.9 & 45.4 & 41.1
\\
CF ~\cite{2020coarse2fine} & ResNet-50 & 49.6 & 49.7 & 62.6 & 23.3 & 43.4 & 27.4 & 30.2 & 44.8 & 41.4
\\
SAP ~\cite{li2020SAP} & ResNet-50 & 49.3 & 49.9 & 62.5 & 23.0 & 44.1 & 29.4 & 31.3 & 45.8 & 41.9
\\
{SFA}~\cite{wang2021exploring}& {ResNet-50} &{46.5}& {48.6}& {62.6}& {25.1}& {\textbf{46.2}}&{ 29.4}&{28.3}&{ 44.0}& {41.3}\\
MTTrans~\cite{yu2022mttrans} &ResNet-50 &47.7& 49.9& 65.2& 25.8& 45.9& 33.8& \textbf{32.6}& 46.5&43.4 \\
\midrule
\textbf{DA-DETR}~& ResNet-50 & \textbf{49.9} &\textbf{ 50.0} & \textbf{63.1} & 24.0 & {45.8} & \textbf{37.5} & {31.6} & \textbf{46.3} & \textbf{43.5}
\\
\bottomrule
\end{tabular}
\end{footnotesize}
\caption{Experimental results (\%) of the scenario Normal weather to Foggy weather: Cityscapes $\rightarrow$ Foggy Cityscapes.}
\label{table:city2fog}
\end{table*}

\section{Experiments}
\label{experi}

This section presents experimentation including experiment setups, implementation details, ablation studies, comparisons with the state-of-the-art and discussion. More details are to be described in the ensuing subsections.

\begin{table}[t]
\centering
\begin{footnotesize}
\begin{tabular}{c|c|c}
\toprule
\multicolumn{3}{c}{\textbf{SIM 10k $\rightarrow$ Cityscapes}}
\\
\midrule
Method & \multicolumn{1}{c|}{Backbone} & \multicolumn{1}{c}{mAP on Car}
\\
\midrule
Deformable-DETR~\cite{zhu2020deformable}  & ResNet-50 & 47.4 \\
\midrule
DAF ~\cite{chen2018wild}   & ResNet-50 & 49.8 \\
SWDA ~\cite{saito2019strong}  & ResNet-50 & 50.5 \\
{SCL}~\cite{shen2019gradientdetach} & {ResNet-50}&{51.6}\\
{GPA}~\cite{xu2020graph} & {ResNet-50} & {51.3}\\
CRDA ~\cite{xu2020category}  & ResNet-50 & 52.1 \\
CF ~\cite{2020coarse2fine} & ResNet-50 & 52.5 \\
SAP ~\cite{li2020SAP}  & ResNet-50 & 52.3 \\
{SFA}~\cite{wang2021exploring} & {ResNet-50} &{52.6}\\
MTTrans~\cite{yu2022mttrans} & ResNet-50 & \textbf{57.9}\\
\midrule
\textbf{DA-DETR}  & ResNet-50 & {54.7} \\
\bottomrule
\end{tabular}
\end{footnotesize}
\caption{Experimental results (\%) of the scenario Synthetic scene to Real scene: SIM 10k $\rightarrow$ Cityscapes.}
\label{table:sim2city}
\end{table}

\subsection{Experiment Setups}
\label{4_1}

\noindent \textbf{Datasets.} 
Following~\cite{chen2018wild, he2019MAF, xu2020category, saito2019strong, inoue2018weakly, kim2019diversify}, we evaluate DA-DETR under four widely adopted domain adaptation scenarios with eight datasets as listed:
1) \textit{Normal Weather to Foggy Weather} (Cityscapes~\cite{cordts2016cityscapes} $\rightarrow$ Foggy Cityscapes~\cite{sakaridis2018foggy}); 2) \textit{Synthetic Scene to Real Scene} (SIM 10k~\cite{johnson2016sim10k} $\rightarrow$ Cityscapes~\cite{cordts2016cityscapes}); 3) \textit{Cross-camera Adaptation} (KITTI~\cite{geiger2013kitti} $\rightarrow$ Cityscapes~\cite{cordts2016cityscapes}) and 4) \textit{Real-world Images to Artistic Images} (PASCAL VOC~\cite{everingham2015pascal} $\rightarrow$ Clipart1k, Watercolor2k, Comic2k~\cite{inoue2018weakly}).

\begin{table}[!t]
\centering
\begin{footnotesize}
\begin{tabular}{c|c|c}
\toprule
\multicolumn{3}{c}{\textbf{KITTI $\rightarrow$ Cityscapes}}
\\
\midrule
Method & \multicolumn{1}{c|}{Backbone} & \multicolumn{1}{c}{mAP on Car}
\\\midrule
Deformable-DETR~\cite{zhu2020deformable}  & ResNet-50 & 39.5 \\
\midrule
DAF~\cite{chen2018wild}   & ResNet-50 & 43.6 \\
SWDA~\cite{saito2019strong} & ResNet-50 & 44.3 \\
{SCL}~\cite{shen2019gradientdetach}&{ ResNet-50}& {44.5}\\
{GPA}~\cite{xu2020graph}& {ResNet-50}& {43.2}\\
CRDA~\cite{xu2020category}  & ResNet-50 & 44.8 \\
CF~\cite{2020coarse2fine}  & ResNet-50 & 45.2 \\
SAP~\cite{li2020SAP} & ResNet-50 & 46.5 \\
{SFA}~\cite{wang2021exploring}& {ResNet-50} &{46.7}\\
\midrule
\textbf{DA-DETR}  & ResNet-50 & \textbf{48.9} \\
\bottomrule
\end{tabular}
\end{footnotesize}
\caption{Experimental results (\%) of the scenario Cross-camera Adaptation: KITTI $\rightarrow$ Cityscapes.}
\label{table:kitti2city}
\end{table}

\begin{table*}[!t]
\centering
\resizebox{\linewidth}{!}{
\begin{tabular}{c|cccccccccccccccccccc|c}
\toprule
\multicolumn{22}{c}{\textbf{PASCAL VOC $\rightarrow$ Clipart1k}}
\\
\midrule
Method & aero & bcyc. & bird & boat & bott. & bus & car & cat & chair & cow & table & dog & horse & bike & pers. & plant & sheep & sofa & train & tv& mAP\\
\midrule
Deformable-DETR~\cite{zhu2020deformable} & 24.8&50.5&14.0&22.8&11.5&50.7&28.7&3.0&26.5&32.6&22.1&17.4&19.6&73.1&54.2&20.8&11.5&12.6&55.2&30.3 &29.1 \\
DAF~\cite{chen2018wild} &33.5&39.6&24.9& 31.4&19.0&61.8&34.5&11.0&29.2&28.5&22.6&20.9&26.5&61.4&51.6&26.7&8.3&23.1&59.7&39.5 &32.7
\\
SWDA~\cite{saito2019strong}&38.6&\textbf{53.0}&29.4&\textbf{39.5}&{25.2}&\textbf{64.8}&36.9&{21.4}&37.9&39.5&30.7&{28.7}&31.4&\textbf{73.7}&63.4&33.5&15.8&29.2&61.3&41.2& 39.8 \\
{SCL}~\cite{shen2019gradientdetach}&{32.3}&{46.8}&{31.9}&{36.0}&{36.8}&{43.6}&{40.9}&{24.4}&{35.1}&{37.8}&{18.1}&{\textbf{34.9}}&{32.6}&{67.3}&{64.5}&{\textbf{43.2}}&{14.5}&{30.4}&{53.5}&{43.6} &{38.4}\\
{GAP}~\cite{xu2020graph} &{28.9} & {42.4} & {32.4} & {36.8}& {36.5}& {40.8}& {39.1}& {23.2}& {34.6} & {39.1}& {16.6} & {33.1} &{ \textbf{36.4}} & {65.2} & {66.0} & {40.1}& {14.3} & {30.6} & {56.4} &{39.5} & {37.6}\\
{SFA}~\cite{wang2021exploring} &{35.2} &{47.6} &{ \textbf{33.5}}&{38.3}&{\textbf{39.6}}&{40.4}&{38.5}&{\textbf{27.2}} &{37.6}&{43.1}& {23.9} &{31.6} &{32.5} &{72.5} &{\textbf{66.8}}&{43.0} &{\textbf{18.5}} &{29.0}&{53.0}& {44.9}&{39.8}\\
\midrule
\textbf{DA-DETR} & \textbf{43.1}&47.7&{31.5}&33.7&21.4&62.8&\textbf{42.6}&14.8&\textbf{39.5}&\textbf{44.2}&\textbf{35.9}&27.5&{31.8}&72.6&{65.6}&{42.2}&17.3&\textbf{31.1}&\textbf{71.3}&\textbf{50.1}& \textbf{41.3}
\\
\bottomrule
\end{tabular}
}
\caption{Experimental results (\%) of the scenario Real-world images to Clipart-style images: PASCAL VOC $\rightarrow$ Clipart1k.}
\label{table:pascal2clipart}
\end{table*}

\subsection{Implementation Details}
\label{4_2}

In all experiments, we adopt deformable-DETR \cite{zhu2020deformable} as the base detector. Since there is only few prior study~\cite{wang2021exploring} on transformer-based domain adaptive detection, we modify existing object detectors using Faster R-CNN~\cite{chen2018wild,saito2019strong,xu2020category, 2020coarse2fine,li2020SAP, guan2021uncertainty} to the transformer-based domain adaptive detection for fair comparisons. 
The modification is accomplished by keeping domain adaptation modules unchanged but replacing post-processing modules in Faster R-CNN ($e.g.$, region proposal network, proposal classification module, etc.) by the encoder-decoder module of deformable DETR. 
In addition, we adopt ResNet-50~\cite{he2016resnet} (pre-trained on ImageNet~\cite{deng2009imagenet}) as backbone, and use SGD optimizer~\cite{bottou2010large} with a momentum $0.9$ and a weight decay $1e-4$ in all experiments with deformable-DETR.

In all experiments, the weight factor $\lambda$ in Eq.~\ref{10} is fixed at 0.1 and the number of split groups $K$ in SMF is fixed at 32. All the experiments are implemented in Pytorch. For evaluation metrics, we report average precision (AP) for each object category and mean average precision (mAP) of all object categories with a threshold of intersection over union (IoU) at 0.5 as in ~\cite{chen2018wild,saito2019strong,xu2020category}.

\begin{table}[t]
\centering
\resizebox{\linewidth}{!}{
\begin{tabular}{c|cccccc|c}
\toprule
\multicolumn{8}{c}{\textbf{PASCAL VOC $\rightarrow$ Watercolor2k}}
\\
\midrule
Method &bike& bird &car &cat &dog& person & mAP \\
\midrule
Deformable-DETR~\cite{zhu2020deformable}&{43.3}&{39.9}&{21.0}&{\textbf{50.3}}&{13.7}&{49.1} &{36.2}\\
{DAF}~\cite{chen2018wild}&{58.0}&{41.7}&{30.2}&{32.7}&{34.5}&{66.9} &{44.0}
\\
{SWDA}~\cite{saito2019strong}&{\textbf{58.7}}&{\textbf{53.7}}&{25.3}&{40.2}&{32.8}&{70.2}& {46.8}\\
{UaDAN}~\cite{guan2021uncertainty}&{57.2}&{47.8}&{31.0}&{37.8}&{34.9}&{70.3}&{48.2}\\
\midrule
{\textbf{DA-DETR}}&{58.6}&{\textbf{53.7}}&{\textbf{31.9}}&{46.2}&{\textbf{40.2}}&{\textbf{73.0}}& {\textbf{50.6}}
\\
\midrule
\multicolumn{8}{c}{\textbf{PASCAL VOC $\rightarrow$ Comic2k}}
\\
\midrule
Method & bike& bird &car &cat &dog& person & mAP \\
\midrule
Deformable-DETR~\cite{zhu2020deformable} & {22.3}&{13.6}&{19.6}&{16.6}&{18.9}&{33.1}&{20.3}\\
{DAF}~\cite{chen2018wild} &{27.8} &{17.5} &{28.7} &{24.5} &{20.8} &{45.5} & {27.5}
\\
{SWDA}~\cite{saito2019strong}&{36.6}&{12.8}&{\textbf{29.5}}&{16.5}&{\textbf{33.2}}&{61.7} &{31.7}\\
{UaDAN}~\cite{guan2021uncertainty}&{37.3}&{17.3}&{25.3}&\textbf{28.5}&{29.0}&{61.9}&{33.2}\\
\midrule
{\textbf{DA-DETR}}&{\textbf{44.2}}&{\textbf{18.1}}&{25.0}&{{27.7}}&{33.0}&{\textbf{62.4}} &{\textbf{35.1}}
\\
\bottomrule
\end{tabular}
}
\caption{{Experimental results (\%) of the scenarios Real-world images to Watercolor-style images: PASCAL VOC $\rightarrow$ Watercolor2k and Real-world images to Comic-style images: PASCAL VOC $\rightarrow$ Comic2k.}}
\label{table:pascal2water}
\end{table}

\subsection{Ablation Studies}
\label{4_3}

The proposed CTBlender consists of split-merge fusion (SMF) and scale aggregation fusion (SAF). We first study the two fusion modules to examine how they contribute to the overall unsupervised domain adaptive detection performance. Table \ref{ablation_1} shows experimental results over the validation data of Foggy Cityscapes under the adaptation scenario `normal weather to foggy weather'.

As Table \ref{ablation_1} shows, the \textit{Baseline}~\cite{zhu2020deformable} trained using the labeled source data only does not perform well due to domain shifts. The model \textit{Direct-align} aligns CNN features directly via adversarial learning which improves the \textit{Baseline} from 28.5\% to 38.4\% in mAP. 
The proposed \textit{SMF} is evaluated under three settings including with \textit{Splitting} operation, with \textit{Shuffling} operation and with both. It can be observed that \textit{SMF} under all three settings outperforms the \textit{Direct-align} consistently, while the SMF with both \textit{Splitting} and \textit{Shuffling} performs the best.
In addition, including \textit{SAF} alone over the \textit
{Direct-align} improves mAP by 3.3\%. The incorporation of \textit{SMF} and \textit{SAF} achieves the best mAP at 43.5\%, demonstrating that \textit{SMF} and \textit{SAF} are complementary to each other.

\begin{figure*}[!ht]
\begin{tabular}{p{3.cm}<{\centering} p{3.cm}<{\centering} p{3.cm}<{\centering} p{3.cm}<{\centering} p{3.cm}<{\centering}}
\raisebox{-0.5\height}{\includegraphics[width=1.0\linewidth]{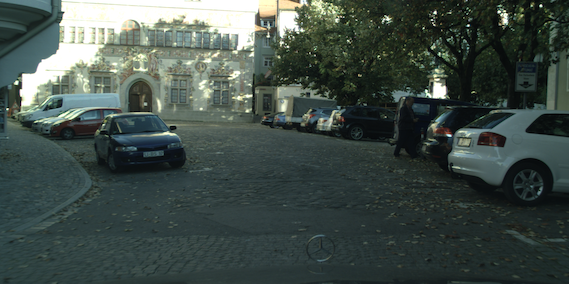}} 
 & \raisebox{-0.5\height}{\includegraphics[width=1.0\linewidth]{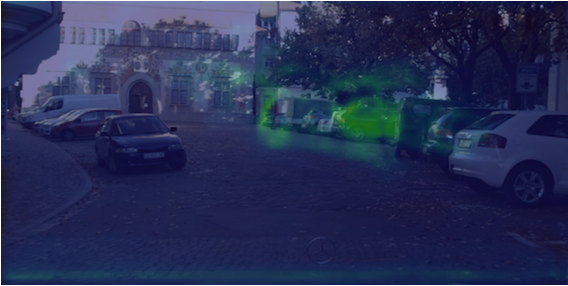}}
& \raisebox{-0.5\height}{\includegraphics[width=1.0\linewidth]{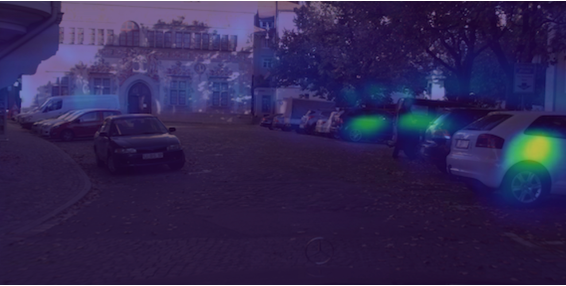}}
& \raisebox{-0.5\height}{\includegraphics[width=1.0\linewidth]{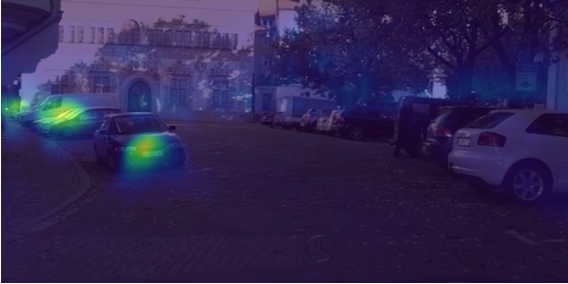}}
& \raisebox{-0.5\height}{\includegraphics[width=1.0\linewidth]{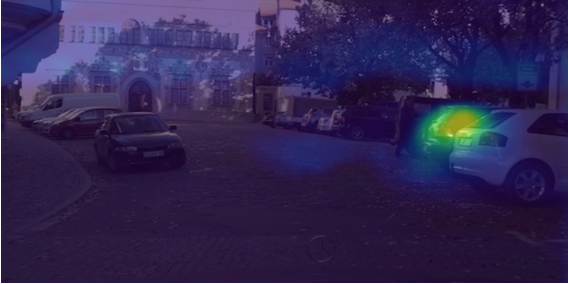}}
\\
\raisebox{-0.5\height}{\includegraphics[width=1.0\linewidth]{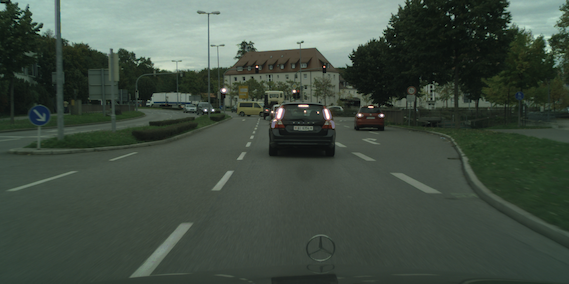}} 
 & \raisebox{-0.5\height}{\includegraphics[width=1.0\linewidth]{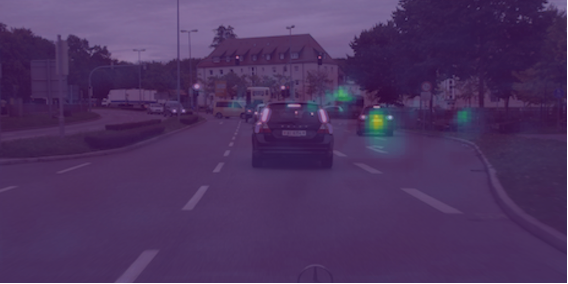}}
& \raisebox{-0.5\height}{\includegraphics[width=1.0\linewidth]{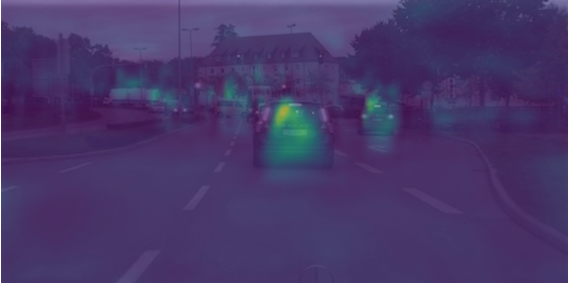}}
& \raisebox{-0.5\height}{\includegraphics[width=1.0\linewidth]{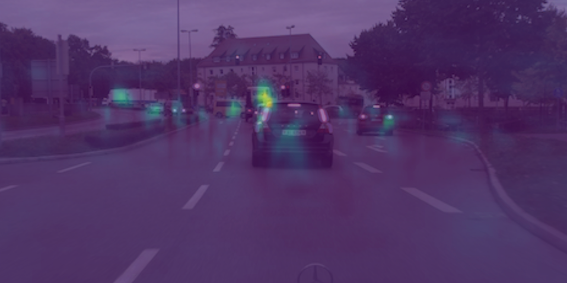}}
& \raisebox{-0.5\height}{\includegraphics[width=1.0\linewidth]{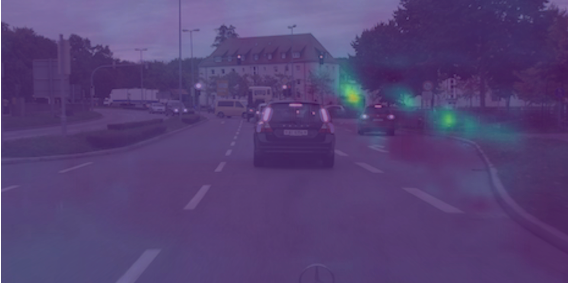}}

\\

\raisebox{-0.5\height}{Original image}
 & \raisebox{-0.5\height}{k=8}
& \raisebox{-0.5\height}{k=16}
& \raisebox{-0.5\height}{k=24}
& \raisebox{-0.5\height}{k=32}
\\
\end{tabular}
\caption{
Visualization of generated weight in SMF for each group. We take two sample images from validation set of Cityscapes, which are shown in the first column. For each image, we sample 4 groups from the total 32 groups and highlight the generated attention over each sample image as shown in columns 2 to 5, respectively. We can observe that the generated weight in different groups detect different foreground regions effectively. $k$ denotes the $k^{th}$ group defined in Section~\ref{method}.
}
\label{fig:group_att}
\end{figure*}

\subsection{Comparisons with the State-of-the-Art}
\label{4_4}

We evaluate DA-DETR under four domain-shift scenarios: 1) Normal weather to Foggy weather; 2) Synthetic scene to Real scene; 3) Cross-camera Adaptations and 4) Real-world images to Artistic images. In each domain-shift scenario, we compare DA-DETR with a number of state-of-the-art unsupervised domain adaptive methods. 

\noindent \textbf{Normal Weather to Foggy Weather:}
We first study the adaptation from normal weather to foggy weather on the task Cityscapes $\rightarrow$ Foggy cityscapes. 
As Table~\ref{table:city2fog} shows, DA-DETR outperforms the baseline deformable DETR~\cite{zhu2020deformable} greatly. It also outperforms the state-of-the-art~\cite{wang2021exploring} by 2.2\% in mAP.
For certain categories such as `train' that can not be well detected by existing methods, DA-DETR achieves the best AP of 37.5. Such experimental results verify that the proposed CTBlender helps to identify both the semantic information and the localization information effectively.

\noindent \textbf{Synthetic Scene to Real Scene:}
Table~\ref{table:sim2city} shows experiments of adaptation from synthetic to real scenes on the task SIM 10k $\rightarrow$ Cityscapes.
We can observe that DA-DETR achieves the best accuracy with a mAP 54.7\%, showing that DA-DETR is powerful when there is only one object category `car' in cross-domain detection task.

\noindent\textbf{Cross-camera Adaptation:}
Table~\ref{table:kitti2city} shows experiments of cross-camera adaptation over the task KITTI $\rightarrow$ Cityscapes. 
We can observe that DA-DETR outperforms the state-of-the-art and improves the baseline model~\cite{zhu2020deformable} from 39.5\% to 48.9\% in mAP. These experiments further show that the proposed DA-DETR can well generalize to different domain adaptation tasks.

\noindent \textbf{Real-world Images to Artistic Images:}
We evaluate the adaptation from real-world images to clipart-style images on the task PASCAL VOC $\rightarrow$ Clipart1k. Table~\ref{table:pascal2clipart} shows experimental results, where DA-DETR achieves the best mAP of 41.3\%. In addition, DA-DETR improves the baseline by large margins for certain categories that are not well detected by the baseline model~\cite{zhu2020deformable} such as bird and sofa. This experiment shows that DA-DETR can handle domain adaptation with multiple categories effectively.

To demonstrate the generalization capability of DA-DETR, we also evaluate it over the tasks PASCAL VOC $\rightarrow$ Watercolor2k and PASCAL VOC $\rightarrow$ Comic2k, respectively. As Table~\ref{table:pascal2water} shows, DA-DETR outperforms the baseline~\cite{zhu2020deformable} by large margins, and it also outperforms all state-of-the-art methods over two tasks consistently.

\subsection{Discussion}
\label{4_5}

\noindent\textbf{Effectiveness of Split Fusion in CTBlender.}
As described in Section~\ref{method}, the split operation in SMF splits input feature into $K$ groups which helps to encode different semantic information into the fused feature. 
We examine the effectiveness of the split operation over domain adaptation task Cityscapes $\rightarrow$ Foggy cityscapes by visualizing the weight generated by each group.
We sampled 4 groups from the total 32 groups for each image and highlighted the produced weight over the sample images as shown in Fig.~\ref{fig:group_att}.
We can observe that SMF captures different foreground regions over different groups, demonstrating that the split operation helps to learn different semantic features effectively.

\begin{table}[t]
\centering
\begin{footnotesize}
\begin{tabular}{c|c|c}
\toprule
\multicolumn{3}{c}{\textbf{Cityscapes $\rightarrow$ Foggy Cityscapes}}
\\
\midrule
Method &\multicolumn{1}{c|}{Aligned Features} & mAP \\
\midrule
Deformable-DETR~\cite{zhu2020deformable} &\multicolumn{1}{c|}{N.A.} & 28.5 \\
\midrule
Direct-Align~\cite{ganin2015grl} &CNN features & 38.4\\
\midrule
Direct-Align~\cite{ganin2015grl} &Transformer features & 38.9\\
\midrule
{Direct-Align}~\cite{ganin2015grl}&&{40.2}\\
{Addition}~\cite{he2016resnet}&& {42.1}\\
{Multiplication}~\cite{woo2018cbam}&\multicolumn{1}{c|}{\multirow{3}{*}{{\begin{tabular}[c]{@{}c@{}}CNN features\\ and \\ Transformer features\end{tabular}}}}&{41.9}\\
{Convolution}~\cite{zhang2020resnest}&&{41.8}\\
{AFF}~\cite{dai2021attentional}&&{42.4}\\
{LS-DeconvNet}~\cite{cheng2017locality}&&{42.6} \\
\textbf{CTBlender(ours)} &\multicolumn{1}{c|}{} & {\textbf{43.5}}\\
\bottomrule
\end{tabular}
\end{footnotesize}
\caption{
Comparing the proposed CTBlender with conventional fusion mechanisms in cross-domain alignment (on the domain adaptive object detection task Cityscapes $\rightarrow$ Foggy Cityscapes).
}
\label{compare_att}
\end{table}

\noindent\textbf{Analysis of CNN Features and Transformer Features.}
We study how CNN features and Transformer features affect unsupervised domain adaptation by examining the adaptation performance of the direct alignment of CNN features $f$, the direct alignment of Transformer features $p$ and both, respectively. 
As shown in Rows 1-4 of Table~\ref{compare_att}, aligning CNN features and Transformer features simultaneously brings clear further performance improvement over either CNN feature alignment or Transformer feature alignment.
This shows that either the localization information in CNN features or the semantic information in Transformer features can facilitate domain adaptation in some degree while these two types of information are complementary for cross-domain alignment.

\noindent\textbf{Comparison with Conventional Fusion Mechanisms.}
We study how different feature fusion strategies affect the domain adaptation performance by comparing our CTBlender with existing feature fusion strategies~\cite{he2016resnet,woo2018cbam,zhang2020resnest,dai2021attentional,cheng2017locality}, $e.g.$, fusing CNN features and Transformer features via 1) addition~\cite{he2016resnet}, multiplication~\cite{woo2018cbam} and convolution~\cite{zhang2020resnest}, and 2) attention-based fusion~\cite{dai2021attentional,cheng2017locality}. As shown in the bottom part of Table~\ref{compare_att}, all fusion strategies improve the \textit{Direct-Align} baseline clearly, demonstrating the effectiveness of aligning the fused features in UDA.
In addition, we can observe that our CTBlender performs the best clearly, largely attributed to its split-merge fusion and scale aggregation fusion designs that fuses CNN features and Transformer features ingeniously.
Specifically, the split-merge fusion in CTBlender splits the CNN/Transformer feature which enables to fuse them along spatial and channel dimensions respectively, leading to comprehensive information fusion along both feature dimensions. Besides, the scale aggregation fusion in CTBlender aggregates rich information across multiple image scales, leading to effective cross-domain feature alignment for different scales that facilitates cross-domain detection against large scale variance.

\section{Conclusion}
\label{conclude}

This paper presents DA-DETR, an unsupervised domain adaptive detection transformer that introduces information fusion into the DETR framework for effective knowledge transfer from a labeled source domain to an unlabeled target domain. We design a novel CNN-Transformer Blender that fuses the CNN features and Transformer features ingeniously for effective feature alignment and domain adaptation across domains.
Extensive experiments over multiple domain adaptation scenarios show that DA-DETR achieves superior performance in unsupervised domain adaptive object detection. 
Moving forwards, we plan to continue to investigate innovative cross-domain alignment strategies for better domain adaptive transformer detection.

\textbf{Acknowledgement.}
This study is supported under the RIE2020 Industry Alignment Fund – Industry Collaboration Projects (IAF-ICP) Funding Initiative, as well as cash and in-kind contribution from the industry partner(s).

{\small
\bibliographystyle{ieee_fullname}
\bibliography{egbib}
}

\end{document}